\documentclass[11pt,a4paper]{article}

\usepackage{amsmath}
\usepackage{amssymb}

\usepackage{graphicx}

\usepackage{color}

\newtheorem{proc}{Procedure}
\newtheorem{proposition}{Proposition}
\newtheorem{theorem}{Theorem}
\newtheorem{remark}{Remark}

\begin{document}

\title{Persistent-homology-based gait recognition\thanks{
Partially supported by MINECO/FEDER-UE under grant MTM2015-67072-P.
General acknowledgments should be placed at the end of the article.
} 
}

\author{J. Lamar-Leon, Raul Alonso-Baryolo , Edel Garcia-Reyes \\
             Advance Technologies Application Center (CENATAV), La Habana, Cuba \\
            $\{$jlamar,rbaryolo,egarcia$\}$@cenatav.co.cu           
          \\
        R. Gonzalez-Diaz \\
        Applied Math Dept. I, IMUS, Universidad de Sevilla, Spain\\          
  rogodi@us.es
}

\date{}
\maketitle  

\begin{abstract}
Gait recognition is  an important biometric technique for video surveillance tasks, due to the advantage of using it at distance. 
In this paper, we present a persistent homology-based method to extract topological features (the so-called {\it topological gait signature}) from the 
the body silhouettes of a gait sequence. 
It has been used before in several conference papers of the same authors for human identification, gender classification,  carried object detection  and monitoring human activities at distance. 
The novelty of this paper is the study of the stability of the topological gait signature under small perturbations and the number of gait cycles  contained in a gait sequence. In other words, we show that the topological gait signature is robust to the presence of  noise in the body silhouettes and to the number of gait cycles contained in a given gait sequence. 
We also show that computing  our topological gait signature of only the lowest fourth part of the body silhouette, we avoid  the upper body
movements  that are unrelated to the natural dynamic of the gait, caused for example by carrying a bag or wearing a coat. 
\end{abstract}

\section{Introduction}

Persons recognition at distance, without the subject cooperation, is an important task in video surveillance. Very few biometric techniques can be used in these scenarios. Gait recognition is a technique with special potential under these circumstances due to its advantages, since the features can be extracted from any viewpoint and at bigger distances than other biometric approaches. 

Currently, there are good results in the state of the art for persons walking under natural conditions (without carrying a bag or wearing a coat). See, for example,  \cite{17,11,10}. 
However, it is not common for people to walk without carrying a bag or anything that changes the natural gait. Moreover, people usually perform movements with the upper body part unrelated to the natural dynamic of the gait.  
Up to now, the most successful approaches in gait recognition  use silhouettes to get the features.
 Among the silhouette-based techniques, the best results have been obtained from the methods based in Gait Energy Images (GEI) \cite{18,17,11,14,16}.  
Besides, the GEI methods have been used to eliminate the effects of carrying a bag or wearing a coat in \cite{1,11,17}. 
Generally, these strategies are affected by a small number of silhouettes (one gait cycle or less). 
Moreover, the temporal order in which silhouettes appear is not captured in those representations, loosing the relative relations of the movements in time. 
Besides, the features extracted by these methods are highly correlated with errors in the segmentation of the silhouettes \cite{3} and these errors frequently appear in the existing algorithms for background segmentation. This implies that GEI
 methods are influenced by the shape of the silhouette instead of the relative positions among the parts of the body while walking. 
The accuracy in gait recognition for persons carrying bag or using coat  can be consulted in \cite{17}
for the CASIA-B gait dataset\footnote{http://www.cbsr.ia.ac.cn/GaitDatasetB-silh.zip}. The authors in \cite{17} used features from the full body but the results were not satisfactory. For instance, the best result for persons walking with coats was $32.7\%$ using lateral view, while the worst result was $24.6\%$, using frontal view. Besides, for persons carrying a bag, the best result was $80.2\%$  using frontal view and the worst $52.0\%$, using lateral view. 

\begin{figure}[ht!]
\centering
\includegraphics[width=2 in]{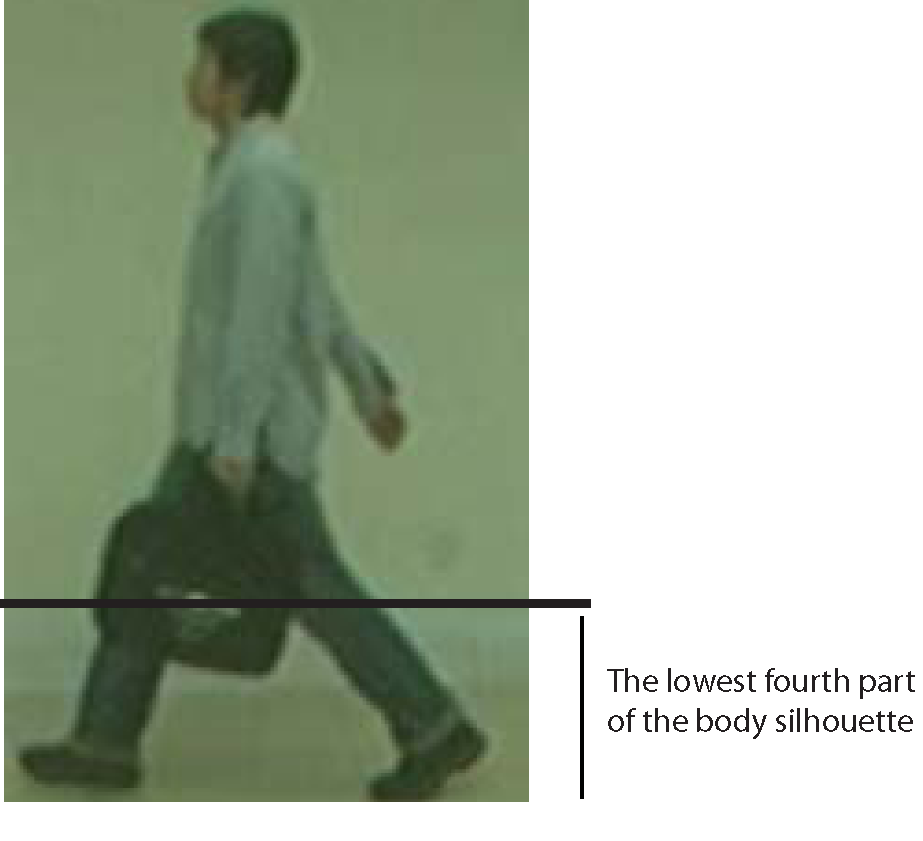} %
\caption{Lowest four part of the body occluded by a bag.}
\label{afect_bag}
\end{figure}

In our  conference papers \cite{10,9,6,7}, 
we concentrated our effort in overcoming most of the difficulties explained above, which took us to get promising results.
In those works, the gait was modeled using a  persistent-homology-based representation (called topological signature of the gait sequence), since it gives features of the objects that are invariant to deformation.  
The topological signature of the gait sequence was used
for human identification in \cite{10}, gender classification
in \cite{9},  carried object detection in \cite{6} and monitoring human activities at distance in \cite{7}.
Later, in \cite{8}, 
we applied our persistent-homology-based gait recognition method using only  the lower part of the body, i.e., the legs (see Fig. \ref{afect_bag}), avoiding many of the effects arising from the variability in the upper body part. 

 In this paper, we first recall 
 our persistent-homology-based  method for gait recognition in order to be self-contained. The input of the  procedure is a sequence of human silhouettes obtained from a video. A simplicial complex $\partial K(I)$
which represents the human gait is then constructed (see Section \ref{simpcompsec}).
Sixteen  persistence barcodes (a known tool in the Theory of Persistent Homology)  are then computed (see  Section \ref{signat}) considering, respectively, the distance to eight fixed planes ($2$ horizontal, $2$ vertical, $2$ oblique and $2$ depth planes) in order to completely capture the movement in the gait sequence. More concretely, for each plane $\pi$, we compute two persistence barcodes: one to detect the variation of connected components and the other to detect the variation of tunnels when we go through  $\partial K(I)$ in a direction  perpendicular to the plane $\pi$.   
Putting together all this information, we construct a vector (called topological signature) associated to each gait sequence. To 
compare two topological signatures, we use the angle between both vectors. 
To decrease the negative effects of variations unrelated to the gait in the upper body part, 
we can only select the lowest fourth part of the body silhouette (legs-silhouette) (as we did in \cite{8}).
 As a contribution of this paper, we  study  the stability of the topological signature in Section \ref{section:stability}. We prove, in terms of probabilities, that small perturbations in the input body silhouettes provoke small perturbations in the resulting topological signature. We also show that the direction of the topological signature (which is a vector) of a gait sequence remains the same independently on the number of gait cycles it contains. To compare two topological signatures, we use the angle between the corresponding vectors, then the previous assertion implies that the topological signature is independent on the number of gait cycles the gait sequence contains. 
Experimental results are showed in Section \ref{exper} and are analyzed in Section \ref{analysis}. Conclusions are finally given in Section \ref{conclusions}. 
%

\section{Topological model of the gait: Simplicial Complexes}
\label{simpcompsec}

In this section we introduce the construction of the simplicial complex $\partial K(I)$ which represents the input human gait sequence.

\begin{figure}[ht!]
\centering
\includegraphics[width=\textwidth]{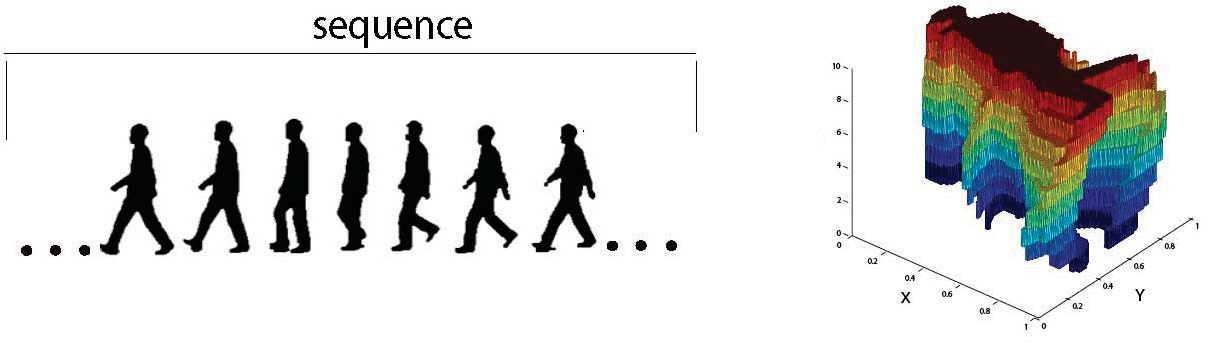} %
\caption{Left: Sequence of gait silhouettes. Right: Simplicial complex $\partial K(I)$.}
\label{fig:dK}
\end{figure}

We start the procedure with a sequence of silhouettes obtained from a gait sequence. With the intention of a fair comparison, we get the sequences from the background segmentation provided in CASIA-B dataset\footnote{http://www.cbsr.ia.ac.cn/GaitDatasetB-silh.zip}. See Fig. \ref{fig:dK}.Left.

We then build a 3D binary  image $I = (\mathbb{Z}^3,B)$  by stacking $k$ consecutive silhouettes. Recall that
 a {\it 3D binary  image}
is a pair  $I=(\mathbb{Z}^3,B)$, where $B$ (called the {\itshape foreground}) is a {\bf finite} subset of $\mathbb{Z}^3$ and  $B^c=\mathbb{Z}^3 \backslash B$ is the {\itshape background}. Later, 
$I=(\mathbb{Z}^3,B)$ is used to derive a cubical complex $Q(I)$.
The {\it cubical complex} $Q(I)$ is a  combinatorial structure constituted by a
set of unit cubes with faces parallel to the coordinate planes and vertices in $\mathbb{Z}^3$, together with all its faces. The $0-${\it faces} of a  cube $c$ are its $8$ corners (vertices), its $1-$faces are its $12$ edges, its $2-$faces are its $6$ squares and, finally, its $3-$face is the cube itself. Then, a  cube with vertices $V = \{(i, j, k), (i+1, j, k), (i, j +1, k), (i, j, k+1), (i+1, j+1, k), (i+1, j, k+1),(i, j+1, k+1), (i+1, j+1, k+1)\}$, with $(i, j, k) \in \mathbb{Z}^3$, is added to $Q(I)$ together with all its faces if and only if $V \subseteq B$.
 Finally, the squares that are faces of exactly one cube in $Q(I)$ are divided into two triangles. These triangles together with their faces (vertices and edges) form the {\it simplicial complex} $\partial K(I)$. 
 See Fig. \ref{fig:dK}.Right.
 The formal definition of  a simplicial complex $K$ is as follows \cite[p. 7]{13}: A simplicial complex $K$ is a collection of simplices\footnote{The simplices considered in this paper are $0-$simplices (i.e. vertices), $1-$simplices (i.e. edges) and $2-$simplices (i.e. triangles).} in $\mathbb{R}^n$ such that: (1) every face of a simplex of $K$ is in $K$; and (2) the intersection of any two simplices of $K$ is a face of each of them. 
Notice that the height of each silhouette is set to $1$ and the   width changes accordingly to preserve the original proportion between height and width. 
$z-$Coordinates represents the amount of silhouettes in the stack which is not fixed (see Fig. \ref{fig:dK}.Right).

\begin{figure}[ht!]
 \centering
 \includegraphics[width=2.5 in]{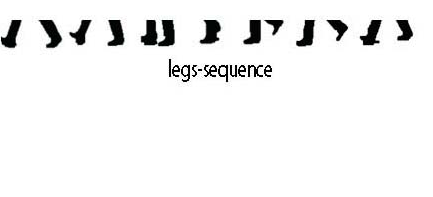} 
 \includegraphics[width=2 in]{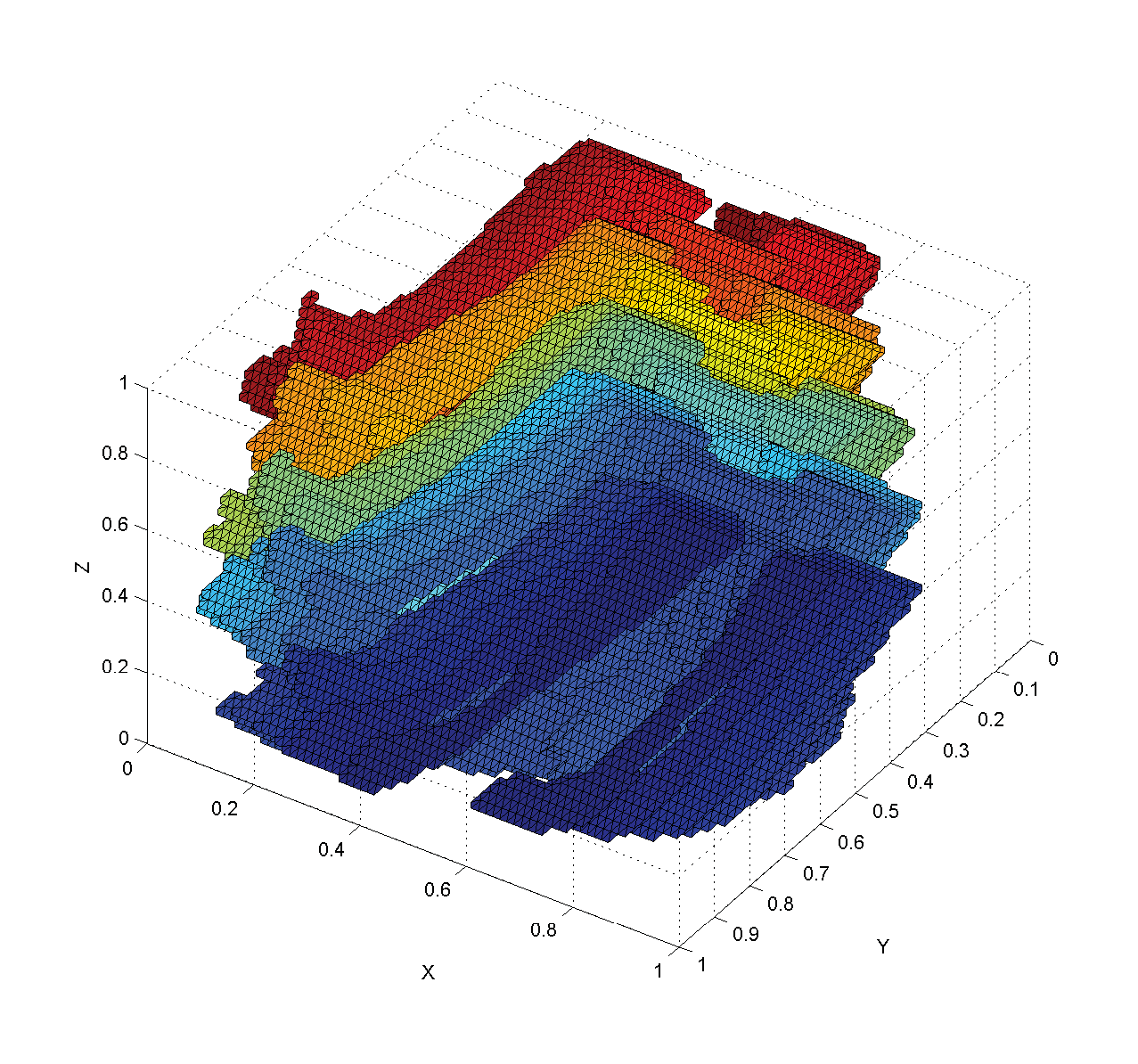} 
 \caption{Left: Sequence of legs-silhouettes. Right: Simplicial complex $\partial K(I)$ of a legs-silhouette sequence.}
 \label{legs}
 \end{figure}

To decrease the negative effects of variations unrelated to the gait in the upper body part (related, for example,  to hand  gestures like talking on cell), in \cite{8} we selected the lowest fourth part of the body silhouette (legs-silhouette) (see Fig. \ref{legs}). This selection is  endorsed by the result given in \cite{1}, which shows that this part of the body provides most of the necessary information for classification.

Finally, notice that, in \cite{8} and this paper, not only the height  but also the  depth is set to $1$. This way, from now on in this paper, $x-$, $y-$ and $z-$coordinates of the vertices in $\partial K(I)$ have their values in the interval $[0,1]$ (see Fig. \ref{legs}.Right).   

\subsection{Filtration of the Simplicial Complex $\partial K(I)$}

The next step in our process is to sort the simplices of $\partial K(I)$ in order to obtain 
a
 filtration.

A {\it filtration}  is a partial ordering of the simplices of $\partial K(I)$
 dictated by a {\it filter function} $f:\partial K(I)\rightarrow \mathbb{R}$, satisfying that if a simplex $\sigma$ is a face of another simplex $\sigma'$ in $\partial K(I)$ then $f(\sigma)\leq f(\sigma')$ (i.e.,  $\sigma$ appears before or at the same time that $\sigma'$ in the  ordering).

\begin{figure}[ht!]
\centering
\includegraphics[width=\textwidth]{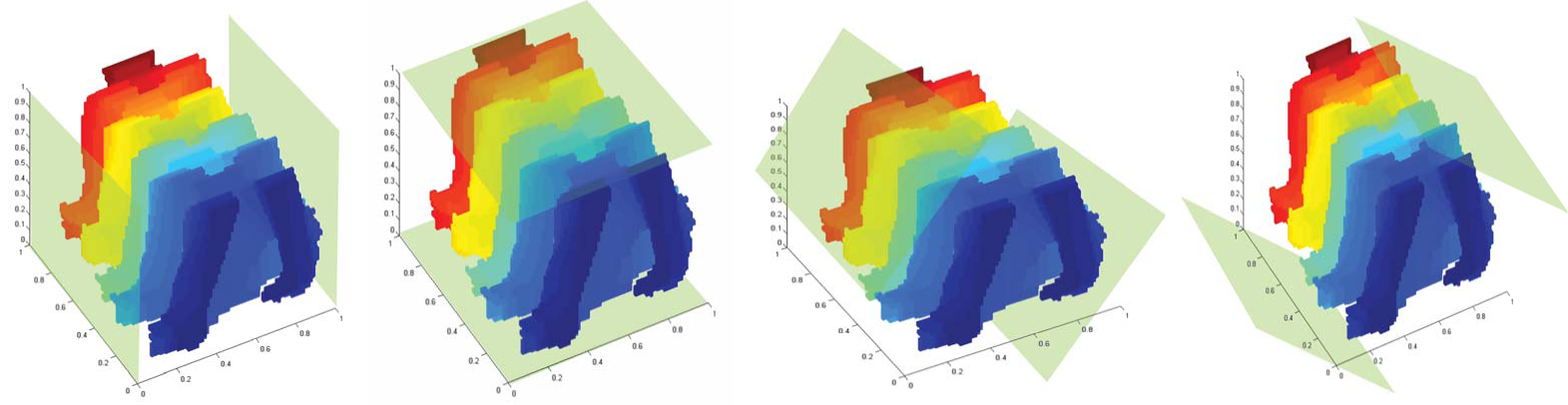} %
\caption{
From left to right: The eight planes
 used for computing the eight filtrations for $\partial K(I)$: two vertical planes, two horizontal planes and four oblique planes. }
\label{planos2_img}
\end{figure}

In our work, we use eight filtrations obtained from eight planes (see Fig. \ref{planos2_img}).
For each plane $\pi$, we define the filter
 function $f_{\pi}:\partial K(I)\rightarrow \mathbb{R}$ which assigns to each vertex of $\partial K(I)$ its distance to the plane $\pi$, and to any other simplex of $\partial K(I)$,  the biggest distance of its vertices to $\pi$. 
Ordering the simplices of $\partial K(I)$ according to the values of $f_{\pi}$, we obtain the filtration $\partial K_{\pi}$ for $\partial K(I)$ associated to the plane $\pi$.

Notice that, in  \cite{10,9,6,7},  the filtration associated to each plane $\pi$ is obtained in a different way: By  adding one simplex at each time (i.e., a total ordering of the simplices is constructed). 
Nevertheless, the filtration presented in \cite{8} and in this paper, is constructed by adding a bunch of simplices at each time (all the simplices of $\partial K(I)$ with same distance to the reference plane $\pi$). This way, different times represent sets of simplices with possibly different cardinalities, which makes the method robust to variation in the amount of simplices of the simplicial complex and therefore, robust to noise. 
Besides, 
the difficulties we had previously  in  \cite{10,9,6,7} with the stability of the sorting algorithm disappear in \cite{8} and in this paper, since  each set of simplices in the filtration contains all the simplices with the same distance to a reference plane, and these sets  are sorted according to their associated distances.

\section{Persistent homology and topological signature} \label{signat}

The final step in our process to obtain the  topological signature of a gait sequence  is to compute the persistent homology of each filtration.

Persistent homology is an algebraic tool for measuring topological features of shapes and functions. It is built on top of homology, which is a topological invariant that captures the amount of connected components, tunnels, cavities and higher-dimensional counterparts of a shape. Small size features in persistent homology are often categorized as noise, while large size features describe topological properties of shapes \cite{4}.

Formally, let $K$ be a simplicial complex. A $p-$chain on $K$ is a formal sum of $p-$simplices of $K$. The group of $p-$chains is denoted by $C_p(K)$. Let us define the homomorphism: $\partial_p:C_p(K)\to C_{p-1}(K)$ called {\it boundary operator} such that for each $p-$simplex $\sigma$ of $K$, $\partial_p(\sigma)$ is the sum of its faces. For example, if $\sigma$ is a triangle, $\partial_2(\sigma)$ is the sum of its edges. The kernel of $\partial_p$ is called the group of {\it $p-$cycles} in $C_p(K)$ and the image of $\partial_{p+1}$ is called the group of {\it $p-$boundaries} in $C_{p}(K)$. The {\it $p-$homology } $H_p(K)$ of $K$ is the quotient group of $p-$cycles relative to $p-$boundaries (see \cite[Chapter 5]{13}). 
Then, $0-$homology classes (i.e. the classes in $H_0(K)$) represent the connected components of $K$, $1-$homology classes  its tunnels  and 
$2-$homology classes its cavities.

Now,  consider a filtration $F$
for  a simplicial complex $K$ obtained from a given filter function 
$f:K\to\mathbb{R}$.
 To simplify the explanation, suppose that the simplices of the filtration are totally ordered (i.e., exactly one simplex is added each time). Let 
  $\ F=(\sigma_1, \sigma_2, \dots, \sigma_m)$.
If $\sigma_{i}$ completes a $p-$cycle ($p$ being the dimension of $\sigma_{i}$) when $\sigma_{i}$ is added to $F_{i-1}=(\sigma_1,\dots,\sigma_{i-1})$, then a $p-$homology class $\alpha$ {\it is born at time $f(\sigma_i)$}; otherwise, a $(p-1)-$homology class {\it dies at time $f(\sigma_i)$}.
 The difference between  the birth and death times of a homology class $\gamma$ is called its {\it persistence}, which quantifies the significance of a topological attribute. If $\alpha$ never
dies, we set its persistence to infinity. 

\begin{figure*}[t!]
\centering
\includegraphics[width=12cm]{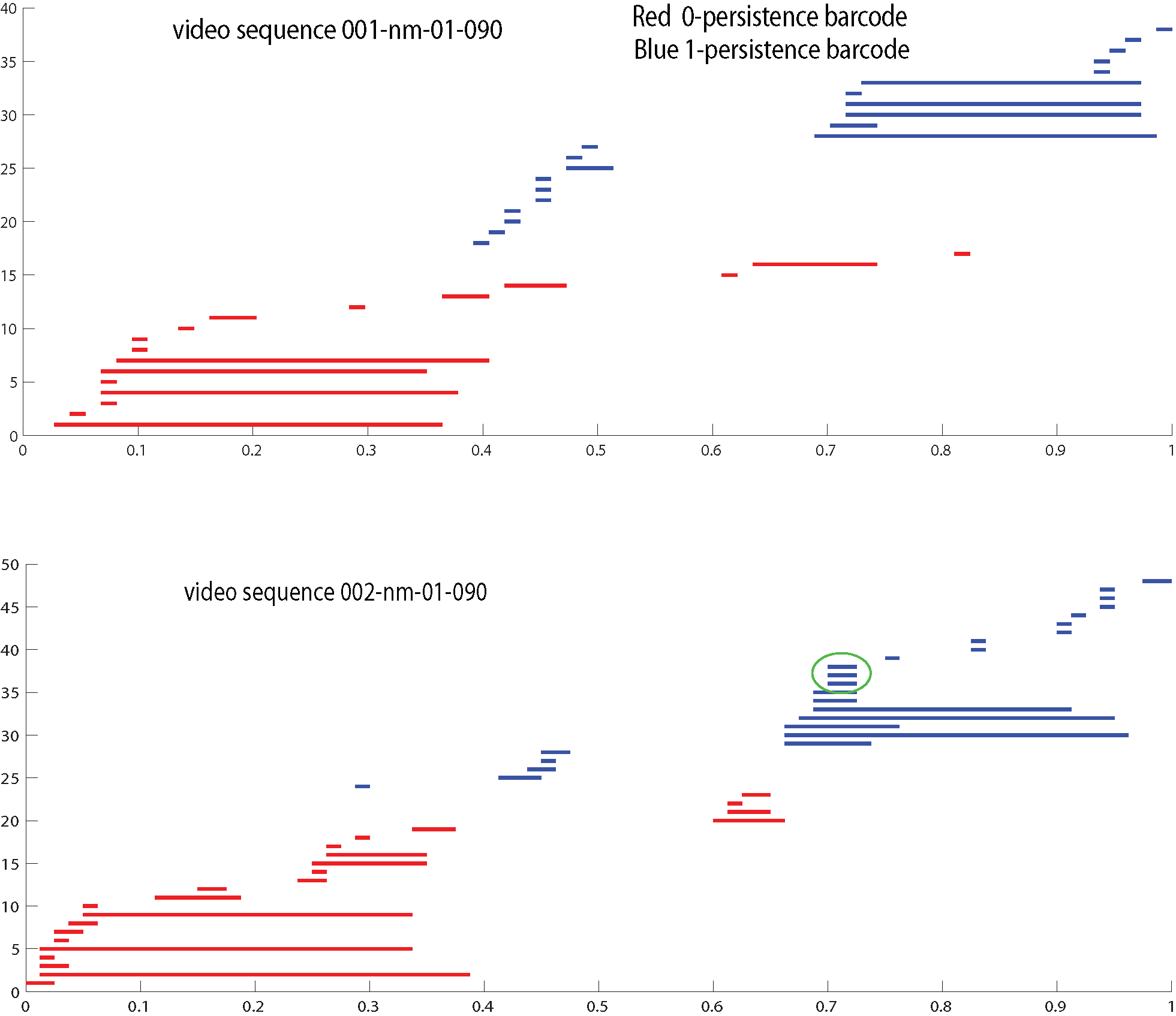} 
\caption{Persistence barcodes for the sequences 001-nm-01-090 and 002-nm-01-090 of CAISA-B dataset. Horizontal axis represents the distance to the reference plane, normalized to $[0.1]$. }
 \label{codebar}
\end{figure*}

For a $p-$homology class that is born at time $f(\sigma_{i})$ and dies at time  $f(\sigma_{j})$, we draw a {\bf bar} 
$[f(\sigma_{i}), f(\sigma_{j}))$
with endpoints  $f(\sigma_{i})$ and $f(\sigma_{j})$.
The set of bars $\{[f(\sigma_{i}), f(\sigma_{j}))\}$ representing birth and death times of homology classes is called the {\it persistence barcode} $B(F)$ of the filtration $F$. See, for example, Fig. \ref{codebar}.
Analogously, the set of points $\{(f(\sigma_{i}), f(\sigma_{j}))\in {\mathbb R}^2\}$  is called the {\it persistence diagram} $dgm(F)$ of the filtration $F$. See, for example, Fig. \ref{diagram}.

\begin{figure}[t!]
\centering
\includegraphics[width=3 in]{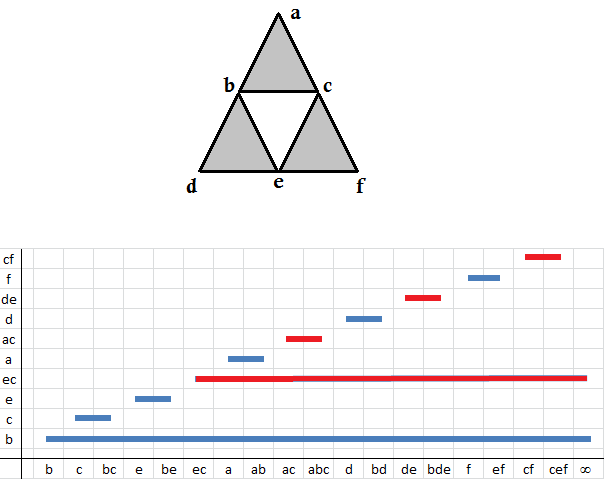} 
\caption{An example of a persistence barcode obtained from a simplicial complex.}
 \label{persistent}
\end{figure}

For example, in Fig. \ref{persistent},  bars corresponding to the persistence of $0-$homology classes (i.e. the persistence of connected components) are colored in blue and bars corresponding to the persistence of $1-$homology classes
(i.e., the persistence of tunnels)
 are colored in red. The filtration 
$F=\{$ $b$, $c$ ,$bc$ ,$e$, $be$, $ec$, $a$, $ab$, $ac$, $abc$, $d$, $bd$, $de$, $bde$, $f$, $ef$, $cf$, $cef\}$
which, in this case, is a total ordering, can also be read on the $x$ axis of the diagram.  Observe that only two bars survive until the end 
(the one corresponding to the connected component and the other corresponding to the tunnel).

For a detailed introduction on the theoretical concepts introduced above see, for example, \cite{4,5}.

As an example, the persistence barcodes for the video sequences \textbf{001-nm-01-090} and \textbf{002-nm-01-090} in the CAISA-B dataset  
are shown in Fig. \ref{codebar}. 
The red bars represent the persistence of $0-$homology classes  while the blue ones represent the persistence of $1-$homology classes.
For computing these barcodes, we used the leftmost plane of Fig. \ref{planos2_img}. 
Notice the green circle showing topological features that born and die at the same time.

\subsection{Topological Signature for a Gait Sequence}


Now, the topological signature  is computed from the persistence barcodes
obtained for 
$\partial K_{\pi}$
 for each plane $\pi$ shown in Fig. \ref{planos2_img}. 
 Observe that fixed a reference plane $\pi$, the length of each interval in the persistence barcode
 obtained for 
$\partial K_{\pi}$ 
is: (a) less or equal than $1$ if $\pi$ is an horizontal o vertical plane, and  (b) less or equal than $\sqrt{2}$ if $\pi$ is an oblique plane.

For computing the topological signature we only consider bars in the persistence barcode with length strictly greater than $0$. This way, we do not take into account any topological event $e$ that is born and dies at the same distance to the reference plane. This is not a problem, since that event $e$ will be capture using a different reference plane. 

Now, for computing the {\it topological signature},  for each plane $\pi$, the $0-$\-persistence barcode (i.e., the lifetime of connected components) and the $1-$\-persistence barcode (i.e., the lifetime of tunnels) of the filtration $\partial K_{\pi}$ are explored according to a uniform sampling. 
More concretely, given a positive integer $n$ (being $n=24$ in our experimental results, obtained by cross validation), we compute the value $h=\frac{k}{n}$, which represents the width of the ``window'' we use to analyze the persistence barcode, being $k$ the biggest distance of a vertex in $\partial K(I)$ to the given plane $\pi$.
Since the distance to the plane has been normalized in order to compute the topological signature, then $k\leq \sqrt{2}$, so $h\leq \frac{\sqrt{2}}{24}$.
Then, 
a vector  ${\cal V}_{\pi}^0$ (resp. ${\cal V}_{\pi}^1$)  of $2n$ entries is then constructed as follows:
\begin{proc}\label{vector}
For $s=0,\dots,n-1$, 
\begin{itemize}
\item[(a)]
entry $2s$ contains 
 the number of $0-$ (resp. $1-$) homology classes that are born before $s\cdot h$; 
 \item[(b)] 
 entry $2s+1$ contains 
 the number of $0-$ (resp. $1-$) homology classes that are born in $s\cdot h$ or later and before $(s+1)\cdot h$.
 \end{itemize}
 \end{proc}

%
%


Observe that dividing the entries in two categories (a) and (b),  small details in the object are highlighted, which is crucial for distinguishing two different gaits. For example,  let us suppose an scenario in which $m$ $0-$homology classes are born in   $[s\cdot h, (s+1)\cdot h)$ and persist or die at the end of $[(s+1)\cdot h, (s+2)\cdot h)$ and not any other $0-$homology class is born, persists or dies in these intervals. Then, we put $0$ in entries $2s$ and $2s+3$ of ${\cal V}_{\pi}^0$, and $m$ in entries $2s+1$ and  $2s+2$ of ${\cal V}_{\pi}^0$. On the other hand, let us suppose that $m$ $0-$homology classes are born and die in   $[s\cdot h, (s+1)\cdot h)$ and in $[(s+1)\cdot h, (s+2)\cdot h)$ and not any other $0-$homology class is born, persists or dies in these intervals. Then, we put $0$ in entries $2s$ and $2s+2$ of ${\cal V}_{\pi}^0$ and $m$ in entries $2s+1$ and $2s+3$ of ${\cal V}_{\pi}^0$. Observe that only considering (a) and (b) separately, we can distinguish both scenarios.
 
This way, fixed a plane $\pi$, we obtain two $2n-$dimensional vectors for $\partial K_{\pi}$, one for the $0-$persistence barcode and the other for the $1-$persistence barcode associated to the filtration  
$\partial K_{\pi}$.
Since we have eight planes, $\{\pi_1,\dots \pi_8\}$, and two vectors per plane,  $\{{\cal V}^0_{\pi_i}, {\cal V}^1_{\pi_i}\}:$ $i=1,\dots,8$, we have a total of sixteen $2n-$dimensional vectors which form the {\it topological signature for a gait sequence}.

Finally, for comparing the topological signatures of two gait sequences, we add up the angle between each pair of the corresponding vectors conforming the topological signatures. Since a signature consists of sixteen vectors, the best comparison for two sequences is obtained when the total sum is zero and the worst is $90\cdot16=1440$. Observe that in our previous papers \cite{10,9,6,7}, we used the cosine distance to compare two given topological signatures. We have noticed that using the angle instead of the cosine, the efficiency (accuracy) increases by $5\%$. This comparison is made in Table \ref{table1} in Section \ref{exper}.

 \section{Stability of the topological signature for a gait sequence}
\label{section:stability}

Once we have defined the {\it topological signature for a gait sequence}, our aim in this section is to prove its stability under small perturbations on the gait sequence and/or  on the number of gait cycles contained
in the gait sequence.

\begin{figure*}[h!]
\centering
\includegraphics[width=\textwidth]{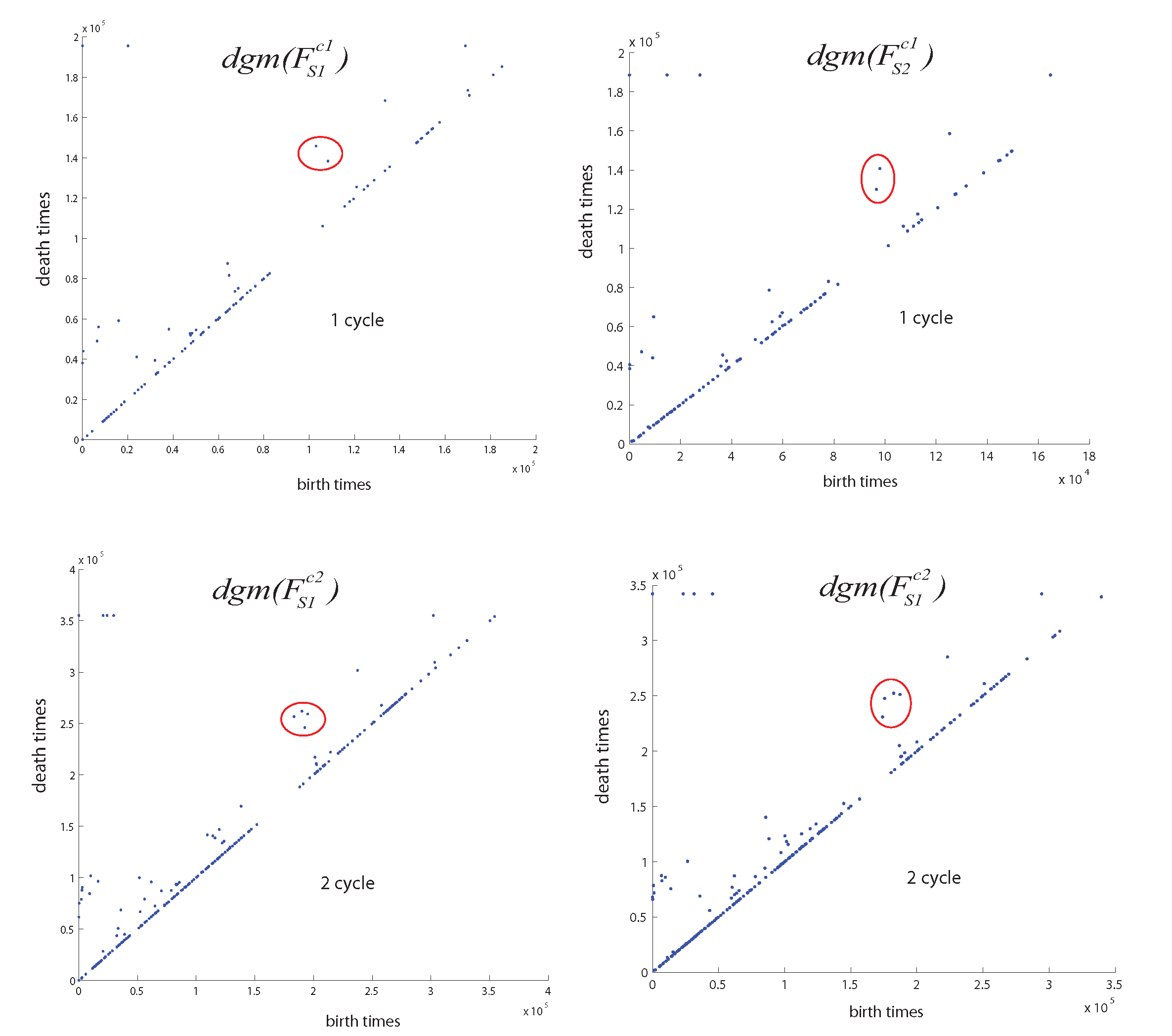} 
\caption{Top pictures: the two $0-$persistence diagrams $dgm(F_{s1}^{c1})$ and $dgm(F_{s2}^{c1})$ obtained from two gait sequences
of the same person containing exactly one gait cycle. Bottom pictures: the $0-$persistence diagrams $dgm(F_{s1}^{c2})$ and $dgm(F_{s2}^{c2})$ obtained from two gait sequences
of the same person containing   two gait cycles. }
 \label{diagram}
\end{figure*}

First, we introduce some theoretical concepts needed to prove the statements above.
The bottleneck distance (see \cite[page 229]{4}) is classically used to compare the persistence diagrams of two different filtrations. 
Concretely, let $F$ and $F'$ be two filtrations of, respectively, two finite simplicial complexes $K$ and $K'$. Since $K$ and $K'$ are finite then 
$dgm(F)$ and $dgm(F')$ are finite.
Let $dgm(F)=\{a_1,\dots,a_k\}$ and $dgm(F')=\{a'_1,\dots,a'_{k'}\}$ be
the persistence diagrams of  $F$ and $F'$,  respectively. Then 
$$d_b(dgm(F),dgm(F'))=\min_{\gamma}\{\max_a\{||a-\gamma(a)||_{\infty}\}\}$$ 
is the {\it bottleneck distance} between $dgm(F)$ and $dgm(F')$
where,
for points $a=(x,y)\in F\cup D$ and $\gamma(a)=(x',y')\in F'\cup D$ (being $D$  the set of points $\{(x,x)\}\subset {\mathbb R}^2$),
$||a-\gamma(a)||_{\infty}=\max\{|x-x'|,|y-y'|\}$ and 
$\gamma$ is a bijection that can associate a point off the diagonal with another point on or off the diagonal. Here, {\it diagonal} is the set $D$. Table \ref{botleneck} shows bottleneck distance between the persistence diagrams shown in Fig. \ref{diagram}.

The following definitions are taken from \cite{2}.
Let $W$ and $W'$ be the   
vertex set of, respectively,  $K$ and $K'$. A {\it correspondence} 
$C:W \Rightarrow W'$ from $W$ to $W'$ is a subset $C$ of $W\times W'$ 
satisfying that  for any $v\in W$ there exists $v'\in W'$ such that $(v,v')\in C$ and, conversely,  for any $v'\in W'$ there exists $v\in W$ such that $(v,v')\in C$. 

Besides, for a subset  $\sigma$ of $W$, $C(\sigma)$ is the subset of $W'$ satisfying that a vertex $v'$ is in $C(\sigma)$ if and only if there exists a vertex $v\in \sigma$ such that $(v,v')\in C$.

Now, given filter functions $f : K\to \mathbb{R}$
and $f': K'\to \mathbb{R}$, and the corresponding filtrations $F$ and $F'$, we say that
$C:W \Rightarrow W'$ is {\it $\epsilon-$simplicial}  from $F$ to $F'$ if for any $t\in \mathbb{R}$ and  simplex $\sigma$ such that $f(\sigma)\leq  t$, every simplex $\mu\in K'$ with vertices in $C(\sigma)$ satisfies that $f'(\mu)\leq t+\epsilon$.

\begin{proposition} \label{prop:previo}
If the correspondence  $C:W\Rightarrow 
W'$ is $\epsilon-$simplicial from $F$ to $F'$,  then 
$d_b(dgm(F),dgm(F'))\leq \epsilon$.
\end{proposition}

\noindent{\bf Proof. }
The statement is a direct consequence of Th. 2.3 and Prop. 4.2 in \cite{2}.
\hfill{$\Box$}

\begin{proposition} \label{prop:previo2}
 Let $I$ (resp. $I'$) be a 3D binary image. 
Let $\partial K_{\pi}$ (resp.  $\partial K'_{\pi}$) be the  filtrations for
$\partial K(I)$ (resp.  $\partial K(I')$)
associated to a given plane $\pi$.
If  $C:W\Rightarrow W'$ is a correspondence from the vertex set $W$ of $\partial K(I)$ to the vertex set $W'$ of $\partial K(I')$ satisfying that 
$f_{\pi}(v')\leq f_{\pi}(v)+\epsilon$ 
 for every $(v,v')\in C$, 
then 
$$\mbox{$C:W\Rightarrow W'$ is   $\epsilon-$simplicial from $\partial K_{\pi}$ to $\partial K'_{\pi}$.}$$
\end{proposition}

\noindent{\bf Proof. }
Let $t\in \mathbb{R}$ and  $\sigma\in \partial K(I)$ such that $f_{\pi}(\sigma)\leq  t$. Then, every simplex $\mu\in K'$ with vertices in $C(\sigma)$ satisfies that $f_{\pi}(\mu)\leq t+\epsilon$.
Then,   $C:W\Rightarrow W'$ is $\epsilon-$simplicial from $\partial K_{\pi}$ to $\partial K'_{\pi}$.
\hfill{$\Box$}

The following is the main result of the paper showing, in terms of probabilities, that the topological gait signature is stable under small perturbations on the input data (i.e., the input gait sequence).

\begin{theorem} 
Let $I$ (resp. $I'$) be a 3D binary image. 
Let $\partial K_{\pi}$ (resp.  $\partial K'_{\pi}$) be the  filtration for
$\partial K(I)$ (resp.  $\partial K(I')$), 
associated to a given plane $\pi$.
Let ${\cal V}^0_{\pi}, {\cal V}^1_{\pi}$ (resp.  
 ${\cal W}^0_{\pi}, {\cal W}^1_{\pi}$) be the 
two vectors obtained after applying Proc. \ref{vector} to the persistence barcodes of $\partial K_{\pi}$ (resp.  $\partial K'_{\pi}$).
Suppose that $m_i=m_i^{K_{pi}}$ is less or equal than $ m_i^{K'_{\pi}}$, where
$m_i^{K_{\pi}}$ (resp. $m_i^{K'_{\pi}}$) is the  number of bars  in the $i-$persistence barcodes (for $i=0,1$) of the filtrations  $\partial K_{\pi}$
(resp.  $\partial K'_{\pi}$).
If  $C:W\Rightarrow W'$ is a correspondence from the vertex set $W$ of $\partial K(I)$ to the vertex set $W'$ of $\partial K(I')$ satisfying that 
$f_{\pi}(v')\leq f_{\pi}(v)+\epsilon$ 
then 
$$\mbox{${\cal V}^i_{\pi} = {\cal W}^i_{\pi}$
with probability greater or equal than $\left(1-\frac{2(n-1)\epsilon}{k}\right)^{m_i}$}$$
where:
\begin{itemize}
\item $k$ is the maximum  distance of a point in $\partial K_{\pi}$ to the plane $\pi$;
\item $n$ is the number of subintervals (``windows") in which the interval $[0,k]$ is divided (recall that we set $n=24$ in our experiments);
\end{itemize}

In general, 
 $||{\cal V}^i_{\pi} - {\cal W}^i_{\pi}||_1\leq m$
with probability greater or equal than 
$$P=\sum_{j=0}^m \left(\begin{array}{c}m_i\\j\end{array}\right)\left(\frac{2(n-1)\epsilon}{k}\right)^j\left(1-\frac{2(n-1)\epsilon}{k}\right)^{m_i-j}.$$
\end{theorem}
Observe that the result above only makes sense when $\epsilon$ is small enough. Concretely, $\epsilon\leq \frac{k}{2n}$ since $\frac{k}{2n}$ is half of the ``window'' size. Besides $m$ can take
 integer values between $0$ and $m_i$.
 Observe that if $m=0$ then $P=\left(1-\frac{2(n-1)\epsilon}{k}\right)^{m_i}$ and if   $m=m_i$ then $P=1$. 

\noindent{\bf Proof. }
By Prop. \ref{prop:previo2},  we have that  $C:W\Rightarrow W'$ is $\epsilon-$simplicial from $\partial K_{\pi}$ to  $\partial K'_{\pi}$.
By Prop. \ref{prop:previo}, we have that 
$d_b(dgm(\partial K_{\pi}),dgm(\partial K'_{\pi}))\leq \epsilon$.
\newline
Let $\gamma: dgm(\partial K_{\pi})\cup\{(x,x)\}\to dgm(\partial K'_{\pi})\cup\{(x,x)\}$ be the bijection such that 
$\{\max_a\{||a-\gamma(a)||_{\infty}\}\}=d_b(dgm(\partial K_{\pi}),dgm(\partial K'_{\pi}))\leq \epsilon$.
If $a=(x,y)$ and $\gamma(a)=(x',y')$, we have that $|x-x'|\leq \epsilon$ and $|y-y'|\leq \epsilon$.
\newline
Now, 
observe that  ${\cal V}^i_{\pi}$ can be different from ${\cal W}^i_{\pi}$ if there exists a point $a=(x,y)$ in $dgm(\partial K_{\pi})$ satisfying that $x\in (sh-\epsilon,sh+\epsilon)$ for $h=\lfloor\frac{k}{n}\rfloor$ and some $s=1,\dots,n-1$. This can occur with probability $\frac{2(n-1)\epsilon}{k}$.
Then ${\cal V}^i_{\pi}={\cal W}^i_{\pi}$
 with probability greater or equal than $(1-\frac{2(n-1)\epsilon}{k})^{m_i}$.
\hfill{$\Box$}

 For example, if $k=0.9$, $n=24$, $\epsilon=0.001$, $m_0=50$
 and $m=3$ then $0.7235226992$; if
 $m=4$ then $P=0.8732685424$ and if $m=5$ then $P=0.9508832893$. This means that  given two digital images $I$ and $I'$ and a plane $\pi$ such that there exists a correspondence $C$ between the vertices of $\partial K(I)$ and $\partial K(I')$ satisfying that 
$f_{\pi}(v')\leq f_{\pi}(v)+0.001$ for any pair of  vertices $v\in\partial K(I)$ and $v'\in partial K(I')$ matched by $C$, then for the associated topological signatures ${\cal V}_{\pi}^0$ and ${\cal W}_{\pi}^0$ (for $k=0.9$, $n=24$ and  $m_0=50$) we have that:
\begin{itemize}
\item $||{\cal V}_{\pi}^0- {\cal W}_{\pi}^0||_1\leq 3$ with probability greater or equal than $72\%$.
\item $||{\cal V}_{\pi}^0- {\cal W}_{\pi}^0||_1\leq 4$ with probability greater or equal than $87\%$.
\item $||{\cal V}_{\pi}^0- {\cal W}_{\pi}^0||_1\leq 5$ with probability greater or equal than $95\%$.
\end{itemize}

The following remark shows that the topological gait signature does not depend on the number of gait cycles in a gait sequence.

\begin{remark}
In the case the gait sequence contains more than a gait cycle, the module of the vectors $\{{\cal V}^0_{\pi_i}, {\cal V}^1_{\pi_i}\}_{i=1,\dots,8}$ can increase with respect to the  gait sequence that contains exactly a gait cycle, but the direction remains  the same.
\end{remark}

\begin{figure*}[h!]
\centering
\includegraphics[width=\textwidth]{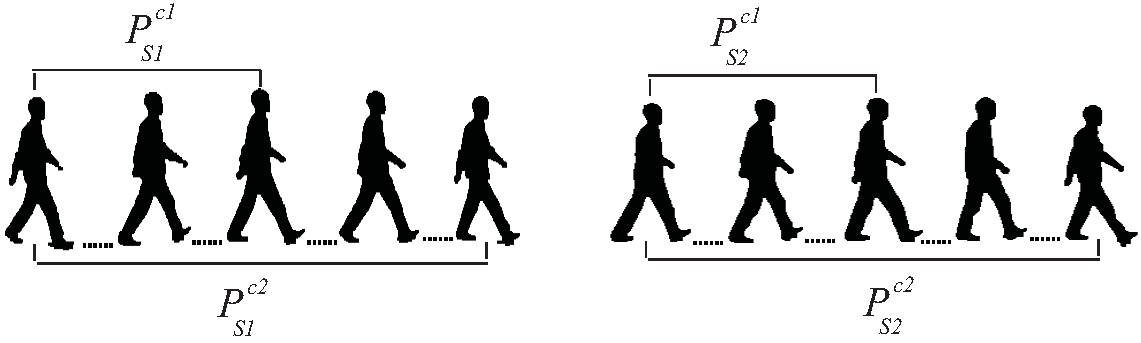} 
\caption{The silhouette sequences  extracted from two gait sequences $S_1$ and $S_2$  of the same person.}
 \label{squence}
\end{figure*}

For instance, let $S1$ and $S2$ be two gait sequences of the same person from CASIA-B dataset, both $S1$ and $S2$ having  two gait cycles. 
Let $P_{s1}^{c2}$ and $P_{s2}^{c2}$ be the silhouette sequences of the two gait cycles on $S1$ and $S2$, respectively.
Let $P_{s1}^{c1}$ and $P_{s2}^{c1}$ be the silhouette sequences of exactly one gait cycle on $S1$ and $S2$, respectively.
See Fig. \ref{squence}.

Then, 
we use the same reference plane $\pi$ to obtain the four filtrations 
$F_{s1}^{c2}$, $F_{s2}^{c2}$, $F_{s1}^{c1}$ and $F_{s2}^{c1}$ for the simplicial complexes associated to the 3D binary images obtained from the  silhouette sequences  $P_{s1}^{c2}$, $P_{s2}^{c2}$, $P_{s1}^{c1}$ and $P_{s1}^{c1}$, respectively.
The $0-$persistence diagrams  $dgm(F_{s1}^{c2})$, $dgm(F_{s2}^{c2})$, $dgm(F_{s1}^{c1})$ and $dgm(F_{s2}^{c1})$ are  showed in Fig. \ref{diagram}. We can observe that the   diagrams 
 $dgm(F_{s1}^{c2})$ and $dgm(F_{s2}^{c2})$ have 
 the double of persistent points than the   diagrams 
 $dgm(F_{s1}^{c1})$ and $dgm(F_{s2}^{c1})$ (look at  the area inside of the red circles  in Fig. \ref{diagram}). Since the topological gait signature is computed using ``windows'' in the persistence barcode (or equivalent, in the persistence diagram), then the modules of the gait signature of $dgm(F_{s1}^{c2})$ and $dgm(F_{s2}^{c2})$ is approximately the double of the modules of the gait signature of $dgm(F_{s1}^{c1})$ and $dgm(F_{s2}^{c1})$ but the direction remains the same.

In Table  \ref{coseno} we show the results of the comparison between the topological signatures obtained from the diagrams 
$dgm(F_{s1}^{c2})$,
 $dgm(F_{s2}^{c2})$, $dgm(F_{s1}^{c1})$ and
 $dgm(F_{s2}^{c1})$
using the  cosine distance. Observe that, in all cases, the cosine distance is almost $1$ (i.e., all the vectors have almost the same direction), which makes sense since all the gait sequences correspond to the same person and the corresponding filtrations are computed using the same reference plane.

\begin{table*}[ht!]
\centering
\caption{Cosine distance between persistence diagrams according to Fig. \ref{diagram}.}
\begin{tabular}{c|c c c }
\hline
 & $dgm(F_{S2}^{c1})$ & $dgm(F_{S2}^{c2})$ \\
\hline
 & &\\
 $dgm(F_{S1}^{c1})$ & 0.985 &  0.987 \\
  & &\\
 $dgm(F_{S1}^{c2})$  & 0.981 &  0.990 \\
\hline
\end{tabular}
\label{coseno}
\end{table*}

Finally, in Table \ref{botleneck} we show that 
 if we consider the classical bottleneck distance for comparing the different persistence diagrams, we obtain different results depending on the number of gait cycles we consider to compute the gait signature. So, the comparison using bottleneck distance does not provide useful information in this case.

 \begin{table*}[ht!]
\centering
\caption{Bottleneck distance between persistence diagrams according to Fig. \ref{diagram}.}
\begin{tabular}{c| c c c }
\hline
 & $dgm(F_{S2}^{c1})$ & $dgm(F_{S2}^{c2})$ \\
\hline
 & &\\
 $dgm(F_{S1}^{c1})$ & 855727 &  1319872 \\
  & &\\
 $dgm(F_{S1}^{c2})$  & 2273559 &  5446584 \\
\hline
\end{tabular}
\label{botleneck}
\end{table*}

We now repeat the experiment for gait sequences obtained from two different persons and {\bf considering only the lowest fourth part of the body silhouettes.} 
Let $001-nm1$ and $001-nm2$ be two gait sequences of the   person $001$ and $008-nm2$ a gait sequence of person $008$ taken from CASIA-B dataset. 


Let $\partial K(I_{001-nm1}^{ci})$,
 $\partial K(I_{001-nm2}^{ci})$ and $\partial K(I_{008-nm2}^{ci})$
 be the simplicial complexes associated to the 
 silhouettes sequences of exactly $i$ gait cycles on $001-nm1$, 
$001-nm2$ and $008-nm2$, for $i=1,2$. 
 
\begin{figure*}[h!]
\centering
\includegraphics[width=\textwidth]{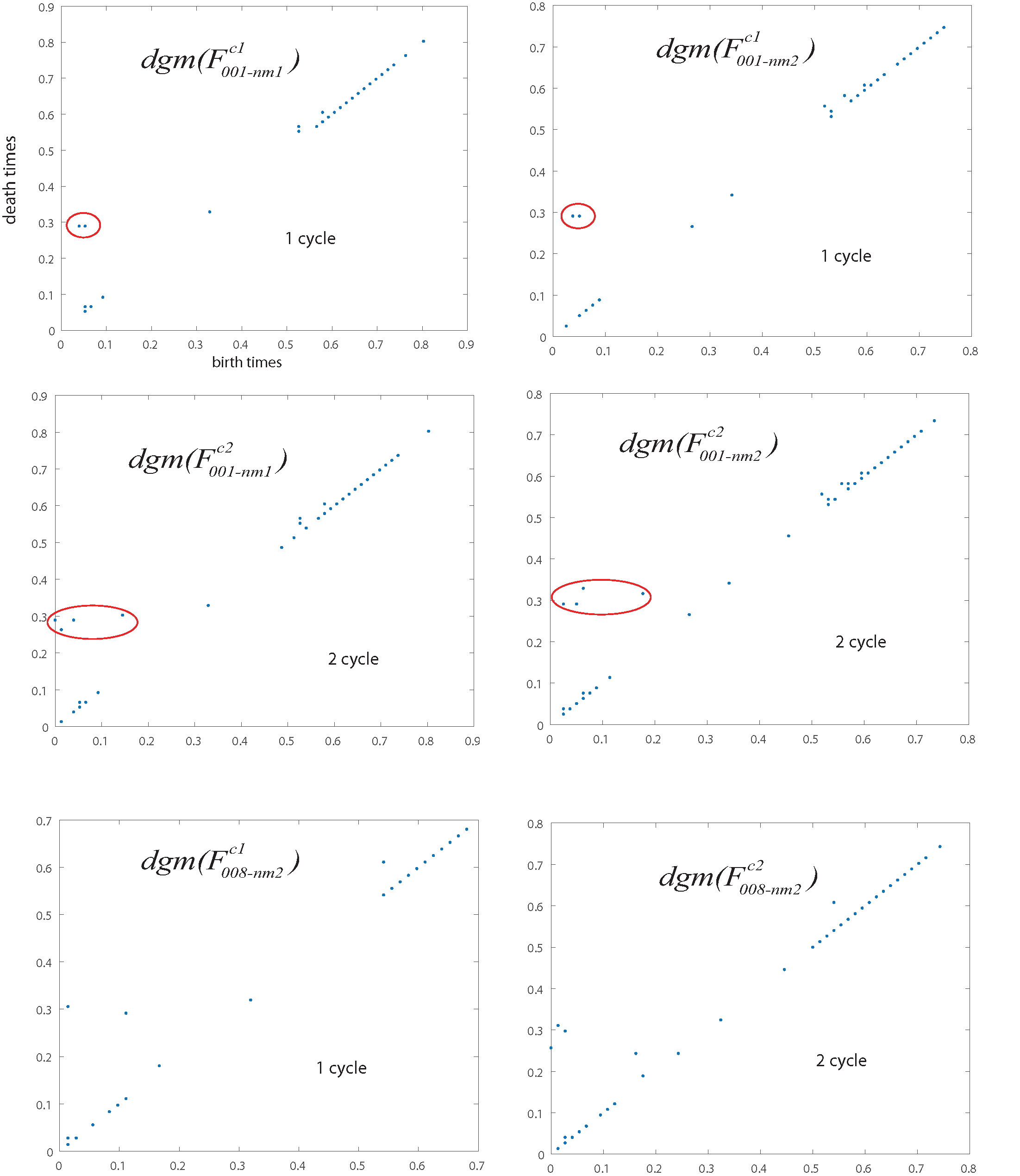} 
\caption{On top, the two $0-$persistence diagrams 
 $dgm(F_{001-nm1}^{c1})$, $dgm(F_{008-nm2}^{c1})$. 
 In the middle, 
 the two $0-$persistence diagrams 
 $dgm(F_{001-nm1}^{c2})$, $dgm(F_{008-nm2}^{c2})$.
 On bottom, 
  the two $0-$persistence diagrams 
 $dgm(F_{008-nm1}^{c1})$, $dgm(F_{008-nm2}^{c2})$.}
 \label{diagram1}
\end{figure*}

A fixed reference plane $\pi$  is used to obtain the $0-$persistence diagrams $dgm(F_{001-nm1}^{ci})$, $dgm(F_{001-nm2}^{ci})$, $dgm(F_{008-nm2}^{ci})$
of the simplicial complex associated to 
$\partial K(I_{001-nm1}^{ci})$,
 $\partial K(I_{001-nm2}^{ci})$ and $\partial K(I_{008-nm2}^{ci})$,
 respectively
(see Fig. \ref{diagram1}). Observe in 
 Fig. \ref{diagram1} that the persistence diagrams obtained from two gait cycles have the double of persistent points than the persistence diagrams obtained from one gait cycle. This can be noticed in the area inside of the red circles  in Fig. \ref{diagram1}.

In Table \ref{botleneck1} and Table \ref{cosine1} we show the results for the comparison between the topological signatures obtained from the previously computed persistence diagrams, using bottleneck and cosine distance, respectively.

 \begin{table*}[ht!]
\centering
\caption{Bottleneck distance between persistence diagrams according to Fig. \ref{diagram}.}
\begin{tabular}{c| c c c c}
\hline
 & $dgm(F_{001-nm2}^{c1})$ & $dgm(F_{001-nm2}^{c2})$ & $dgm(F_{008-nm2}^{c1})$ & $dgm(F_{008-nm2}^{c2})$\\
\hline
 & &\\
 $dgm(F_{001-nm1}^{c1})$ & 0.013 &  0.120 & 0.06 & 0.128\\
  & &\\
 $dgm(F_{001-nm1}^{c2})$  & 0.125 &  0.040 & 0.125 & 0.059\\
\hline
\end{tabular}
\label{botleneck1}
\end{table*}

\begin{table*}[ht!]
\centering
\caption{Cosine distance between the persistence diagrams showed in Fig. \ref{diagram1}.}
\begin{tabular}{c |c c c c}
\hline
 & $dgm(F_{001-nm2}^{c1})$ & $dgm(F_{001-nm2}^{c2})$ & $dgm(F_{008-nm2}^{c1})$ & $dgm(F_{008-nm2}^{c2})$\\
\hline
 & &\\
 $dgm(F_{001-nm1}^{c1})$ & 0.944 & 0.931 & 0.739 & 0.886\\
  & &\\
 $dgm(F_{001-nm1}^{c2})$  & 0.929 &  0.953 & 0.832 & 0.905\\
\hline
\end{tabular}
\label{cosine1}
\end{table*}

The results show that the value of the bottleneck distance increases with the number of gait cycles. However, the values are similar when the cosine distance is used. Furthermore, the comparative between diagrams with two cycles, which mean more information of the dynamic of gait, could improve the similarity, i.e, the bottleneck distance decreases and the cosine distance is closer to one.  

The tables show that the cosine distance is more appropriate than bottleneck distance to compare gait signatures.

\section{Experimental Results}\label{exper}

In this section we show the accuracy results in two experiments using CASIA-B dataset. The CASIA-B dataset has 124 persons, and 10 samples for each of the 11 different angles at which a person is taken. For each angle there are six samples walking under natural conditions, which means without carrying a bag or wearing a coat (CASIA-Bnm), there are two samples of persons carrying some sort of bag (CASIA-Bbg) and the remaining two samples for persons wearing coat (CASIA-Bcl). 
CASIA-B dataset provides image sequences with background segmentation for each person. 

In the first experiment we  used four sequences by person from CASIA-Bnm dataset to train. We used the other two sequences by person from CASIA-Bnm and the sequences from CASIA-Bbg and CASIA-Bcl to test. Our results for lateral view (90 degrees) are shown in Table \ref{table1}, where we take the cross validation average ($\binom {6} {4}$  = 15 combinations) of accuracy at rank 1 from the candidates list. The result of our previous method \cite{10} was also evaluated using always the lowest fourth part of the body silhouette.

\begin{table*}[ht!]
\centering
\caption{Accuracy (in $\%$) using training sets consisting of samples under similar covariate conditions (without carrying a bag or wearing a coat).}
\scalebox{1}{
\begin{tabular}{ccccc}
\hline
Methods & CASIA-Bbg & CASIA-Bcl & CASIA-Bnm & Average \\
\hline

Tieniu.T \cite{17}  & 52.0 & 32.73 & 97.6 & 60.8\\
Khalid.B \cite{1} & 78.3 & 44.0 & 100 & 74.1\\
Singh.S \cite{15}& 74.58 & 77.04 & 93.44 & 81.7\\
Imad.R et al. \cite{14}& 81.70 & 68.80 & 93.60 & 81.40  \\
Lishani et al. \cite{12}& 76.90 & 83.30 & 88.70 & 83.00 \\
\textbf{Previous Method } \cite{10}     & \textbf{75.8} & \textbf{75.45} & \textbf{90.3} & \textbf{80.5}\\
\textbf{Our Method }      & &  &  & \\
using cosine       & \textbf{80.5} & \textbf{81.7} & \textbf{92.4} & \textbf{84.9}\\
using angle       & \textbf{84.2} & \textbf{87.6} & \textbf{94.1} & \textbf{88.6}\\
\hline
\end{tabular}
\label{table1}}
\end{table*}

In the second experiment, we followed the protocol used in \cite[Section 5.3]{1}. This way, we considered a mixture of normal, carrying-bag and wearing-coat sequences, since it models a more realistic situation where persons do not collaborate while the samples are being taken. Specifically,  six sequences were used  to training (four normal sequences, one carrying-bag sequence and one wearing-coat sequence),  the rest was used to test.  Table \ref{table2} shows the result of the accuracy.

\begin{table*}[ht!]
\centering
\caption{Accuracy (in $\%$) using training sets consisting of samples under
different covariate conditions (walking-normal, carrying-bag and wearing-coat).}
\scalebox{1}{
\begin{tabular}{c c c c c}
\hline
Methods & CASIA-Bbg & CASIA-Bcl & CASIA-Bnm & Average\\
\hline
Khalid.B \cite{1} & 55.6 & 34.7 & 69.1 & 53.1\\
\textbf{Our Method}  & \textbf{92.3} & \textbf{94.3} & \textbf{94.7} &  \textbf{93.8}\\
\hline
\end{tabular}
\label{table2}}
\end{table*}

\begin{figure}[h!]
\centering
\includegraphics[width=3.5 in]{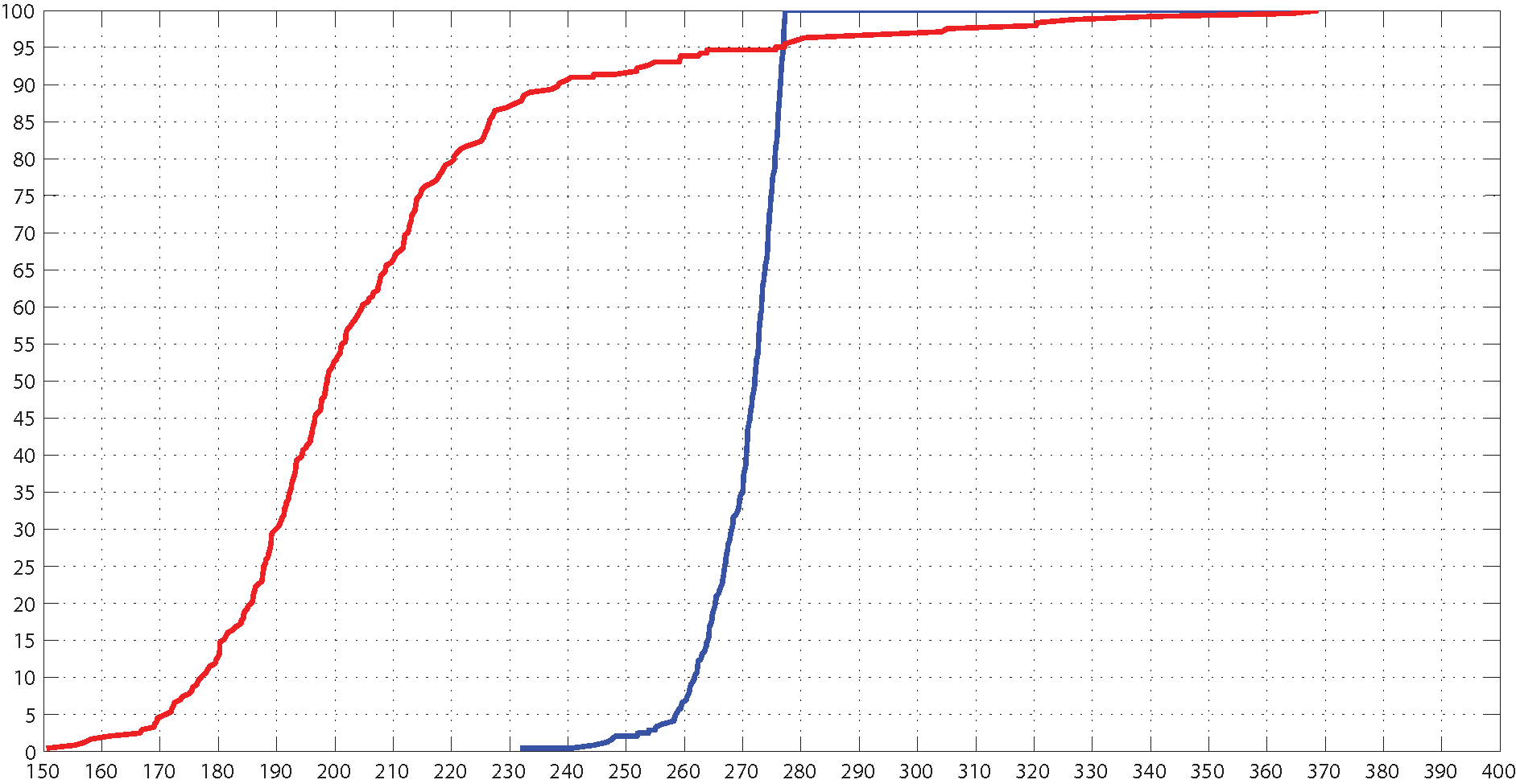} %
\caption{Result of the comparison  between the training set and the test set for the topological signatures of the sequences representing the same person (in red) and the one between the training set and the test set for the topological signatures of the sequences representing different persons (in blue).}
\label{unnamed}
\end{figure}

A third experiment was carried out for obtaining Fig. \ref{unnamed}. In this case six sequences were used for training (one with the person carrying a bag, one with the person wearing a coat and four with the person walking under natural conditions). Using this training data we generated $123$ topological signatures, one for each person in the database. We used the remaining sequence of the person carrying a bag and the one wearing a coat for testing. This gave us $246$ sequences for testing: $123$ persons times $2$ sequences by person. We must point out that person labeled as $005$ in CAISA-B was removed from the experiment due to poor quality.




Using the obtained topological signatures and testing set we obtained:
\begin{itemize}
\item[(1)] The set of all possible comparisons between the obtained signatures and the signatures of the test sequences, corresponding to the same person. This set is called True Positive (TP) and contains $246$ comparison values.

\item[(2)] The set of all possible comparisons between the obtained signatures and the signatures of the test sequences, corresponding to different persons. This set is called True Negative (TN) and contains $123*244=30012$ values.

\end{itemize}

For obtaining Fig. \ref{unnamed}, we first restricted TN to its $246$ smallest values, in order to balance the sets TP and TN. Then we represented in the $y$ axis the percent of values in the TP set lower than a threshold as a red curve and the same for the TN set as a blue curve. The $x$ axis represents all the considered thresholds. For example, $92.6\%$ of the data in the set TP are values smaller than $253.8$, since the red curve shows that for $x=253.8$ we have that $y=92.6\%$.



 
 

More examples and the source code written in Matlab can be obtained visiting the web page: {\it http://grupo.us.es/cimagroup/}.

\section{Analysis of the results}\label{analysis}

In Table \ref{table1} we see that  the best result of our method was for the set of normal sequences (CASIA-Bnm) and the worst was for the set of persons carrying bags.  This is due to that bags can affect the accuracy of the method (see the lowest fourth part of the body silhouette in Fig. \ref{afect_bag}). Moreover, the weight of the bag can change the dynamic of the gait. On the other hand, the features obtained from the lowest fourth part of the body silhouette gave an accuracy for the normal sequences of $94.1\%$, which only decreases $3.9\%$ with respect to our previous paper \cite{10} using the whole body silhouette ($98.0\%$). This confirms that the highest information in the gait is in the motion of the legs, which supports the results given in \cite{1}.

Nevertheless, as we can see in Tables \ref{table1} and \ref{table2},  our method outperforms previous methods for gait recognition with or without carrying a bag or wearing a coat. Furthermore, we show in Table \ref{table1} that the changes introduced to obtain our new method derive in an improvement with respect to our previous solution \cite{10}.

Besides,  the algorithm explained in \cite{1} decreases considerably the accuracy obtained by training mixing   the  normal, carrying-bag and wearing-coat sequences  (see Table \ref{table2}). On the contrary, our algorithm improves the accuracy for the whole test set. Comparing the two tables, we can as well arrive to the conclusion that training with more heterogeneous data gives to our method a more powerful representation for the classification step and in that case, our method outperforms in more that 35\% the results given in  \cite{1}.

Finally, the results presented in this paper  show the power of the Theory of Persistent Homology to obtain the structural features of the dynamic of the legs.

\section{Conclusion}\label{conclusions}
 
In this paper we have presented an algorithm for gait recognition, a technique with special attention in tasks of video surveillance. 
 We have used persistent  homology to model the gait,  similar  as we did in our previous approaches,
although the algorithm presented here is slightly different to our previous ones in a way that it is more robust to variations in the amount of simplices considered
and removes as well any dependency with respect to the stability of the sorting algorithm used for obtaining the filtration. Besides, the topological features have been tested here using only the lowest fourth part of the body silhouette. Then, the effects of variations unrelated to the gait in the upper body part, which are very frequent in real scenarios, decrease considerably. 
We have proved that our topological signature is robust to small perturbations in the input data and does not depend on the number of gait cycles contained in the gait sequence.
 Finally, the results presented in the paper improve considerably the accuracy in the state of the art.

\subsubsection*{Acknowledgments} This work has been partially supported by IMUS and MINECO/FEDER-UE under grant MTM2015-67072-P. 


\end{document}